# A Comprehensive Survey on Aspect Based Sentiment Analysis

Kaustubh Yadav

School of Computer Science and Engineering- SCOPE

Vellore Institute Of Technology, Vellore, Tamil Nadu

**Abstract**

Aspect Based Sentiment Analysis (ABSA) is the sub-field of Natural Language Processing that deals with essentially splitting our data into aspects ad finally extracting the sentiment information. ABSA is known to provide more information about the context than general sentiment analysis. In this study, our aim is to explore the various methodologies practiced while performing ABSA, and providing a comparative study. This survey paper discusses various solutions in-depth and gives a comparison between them. And is conveniently divided into sections to get a holistic view on the process.

I. INTRODUCTION

Aspect-Based Sentiment Analysis can be broken down into two major problem sets, one is Aspect Extraction and the other is Sentiment analysis. Opinion Mining is also a term that is used in place of Aspect Extraction. Aspect Extraction is itself a widely researched area. The exact meaning of Aspect Extraction is the extraction of subjective information from the source material.

Opinion mining on review data can prove to be very beneficial for both the producers and the consumers. In review data sets, ABSA is preferred as instead of giving the overall sentiment it is more precise and biased towards the aspect in hand. An aspect-level opinion Mining decodes, different aspects regarding the item and the polarity towards them. An aspect is defined as an attribute or a feature an item possesses. To perform ABSA, the identification of an aspect is a crucial step. The aspects are of two types, implicit and explicit [1], for example, "The cost of the house is very high, but the house is neat". Here "Cost" is the aspect for the "House" and the polarity is negative towards the house. Also, in the above-given example, the cost is an explicit aspect and neat is implicit. This shows it's not easy to extract implicit and explicit aspects, some models have to use Rule-Based methods[3], based on semantic similarities[4], SVM-Based algorithms[5], Conditional Random Fields (CRF)[6].

Although there are many more methods than mentioned all have some limitations associated with them, for example in order to get precise data from CRF the data set should be of a higher order. Also, some aspects might be tagged incorrectly by the POS tagger[7] as it considers both the noun and noun phrases as an aspect, and all nouns might not be the aspects. A method that first identifies aspects using POS then dependency parsing using CoreNLP and followed by hierarchical clustering has been widely accepted for getting better results[8,9,10,11], this method, when compared with CNN, proved to be better as CNN is known to sometimes fail to identify actual aspect terms. The improved model has a mixture of the Rule-Based and the CNN approach.

After getting the aspect terms next part of per-processing is the generation of Sentiment polarities of the aspect of the item. There are majorly two methods that are covered under this section, aspect-category sentiment analysis (ACSA) and aspect-term sentiment analysis (ATSA)[12]. In ACSA, we calculate sentiment polarity using an aspect which is from a predefined set of categories, in which ATSA sentiment polarity is calculated over a single or a long text. For example "Indian Laptops are very cheap" here "Indian Laptops" is the entity for ATSA.

Many models have been proposed for retrieving sentiment polarity, most of them are based on LSTM layers, through which sentiment information is retrieved from word embedding[14], some models used Attention-based LSTM with aspect embedding for ACSA[15]. And for ATSA, Gated Neural Networks [16,17], Target-Dependent Sentiment Classification (TD-LSTM)[18] and Recurrent Attention Memory Network (RAMN)[19].

After the completion of pre-processing and correct POS tagging, sentiment analysis is performed. In this survey paper, we aim to discuss the complete process from pre-processing to sentiment extraction.

This paper is divided into 3 sections, the first section includes the methods for pre-processing (Section 1), the next sections includes papers which use machine learning algorithms like SVM, (Section 2), the final section of literature include models that involve Neural Networks (Section 3). The Evaluation section includes a comparison between the aforementioned models.

II. GENERIC PRE-PROCESSING FOR ABSA

Pre-processing is very important when we work with any sort of data [23]. There are some standard set of steps but as I read the papers we will discuss these are the steps that were followed.

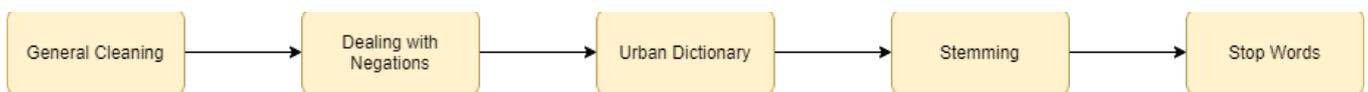

Figure 1: Procedure of generalized Pre-processing

**General Cleaning:** This step includes the removal of disturbing elements and unimportant words for the normalization of misspelled words. Also replacing breaks, commas with spaces.

**Dealing with Negations:** Negations are a crucial part of Sentiment Analysis, ignoring them will give false results. So for uniformity words like (can't, don't, etc) are tagged as not.

**Urban Dictionary:** This step deals with substituting slang with defined words. For example, "h8" is replaced with "hate". When dealing with review data, undefined data such as "coooooool","baddd" are used these words convey meaning, they exaggerate the true meaning hence they are replaced by "very cool" and "very bad", this is done by using a Rule-Based Approach [24].

**Stemming:** This technique deems similar words under a root word, for example, "nearest"," nearer" both are used to project the expression of "near", hence "near" is the root word.

**Stop Words:** These words generally include pronouns, articles, etc. It is crucial to remove these stop words in order to get accurate results.

After considerate research on this, Angiani et.al concluded that Basic processes and Stemming give the best results[25].

Now we move toward aspect extraction, one of the most common methods is POS tagging. Aspects are usually a noun or noun-phrases but this ideology is not uniform[3]. For example, in the "His strides are weak" strides is a verb, and the aspect as well but if we go for generic Opinion Mining it wouldn't recognize any aspect in the aforementioned sentence. So we have to go for different models in order to extract the correct aspect.

As far as POS tagging is concerned it is a widely researched area. Already the accuracy of the Stanford tagger is about 97.3% [20], and Manning[21] gives a model through with it can be boosted up to 100%, but it's important to note that this accuracy is for a token and not for a sentence, if we look at current accuracy for a sentence it is close to 56%. Rush et.al in their paper [22] discuss how consistency constraints with Minimum Spanning Tree (MST) parser and Stanford POS tagger can help significant error reduction in sentence parsing.

As we observed, POS tagging doesn't work well with sentences as it does with actual sentences opposed to tokens. Here is a figure that shows different models for Aspect Extraction.

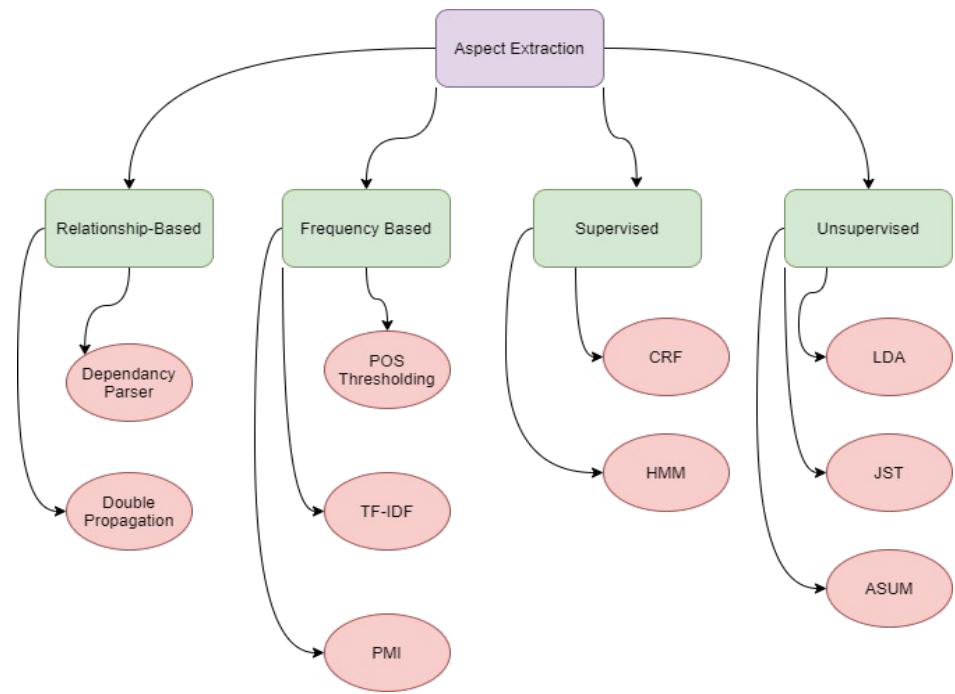

Figure 2: Classification of Aspect Extraction.

**Relationship-Based:**

The method behind this approach is to find relationships between the opinion words and the aspects. This is achieved by looking at grammatical and syntactic patterns. In Peng et al. [26], dependency parsing was described as a method to find a dependance parse that gives a semantic/syntactic relation with the word and the roots. In Zhunag et al. [27] used the IMDB dataset for experimentations, and they used this dependency parsing for understanding the dependency of words/ word phrases with each other. For example, "This movie is a masterpiece" after parsing can be represented as:

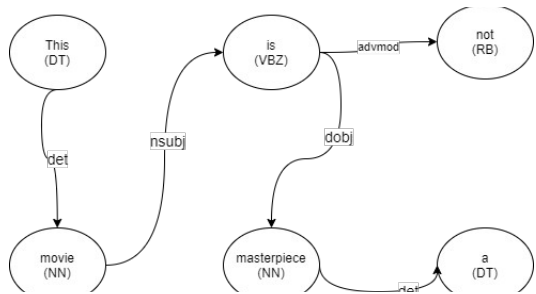

Figure 3: Sentence-Breakdown using Dependancy Parser

Through this parse tree, it can be said derived that "movie" is the aspect and the sentiment is of a positive or negative nature. In Yuanbin et al. [28] phrase dependency parser is used to extract the noun and the word phrases, the generic dependance parser just gives dependency of words or tokens, but this gives dependencies between noun and verb phrases hence improving the overall accuracy for aspect extraction. In Qiu et al.[29], after finding the relations using dependency parsing, which can then be used to expand opinion lexicon and for easier extraction, this is called Double Propagation. The advantage of this method is that it only needs one initial opinion lexicon, so its sort of semi-unsupervised. One limitation of this approach is that it can generate a lot of non-aspects as it looks for syntactic patterns.

**Frequency-Based**

This method is based on calculating the frequency of nouns and noun phrases, in other words it finds the explicit aspects from the data, there are some drawbacks to this model, firstly as we discussed before not every aspect is a noun or a noun-phrase which will generate non-aspects. Secondly, it's easy to miss low-frequency aspects.

As introduced by Kelledy et al. [30], in this method using POS-tagging the count for the noun and the noun phrases is calculated and a threshold value is decided through manual intuition, and frequency lower than that is not considered as an aspect.

As we have seen some, non-nouns might be aspects, it is also true that all nouns or noun-phrases are not aspects, so for precise results we need to remove these words. In Oren et al. [31], the method used first calculates pointwise mutual information(PMI) score between the phrases and meronymy associated with the entity.

Blair et al.[32] used the frequency method but the difference is that they only took into consideration the sentences that have an opinion. Endo et al. [33], used TD-IDF to find frequent terms in their dataset that are related to the aspect in hand. Ester et al. [34] made a significant improvement to the frequency-based approach by removing non-aspect terms using syntactic patterns.

**Supervised:**

Supervised Learning refers to the creation of a model based in a training set which should be labeled and then applying it to a testing set which is unlabelled. So identification of aspects, opinion is annotations problem, where syntactic patterns are labeled data. Hidden Markov Model(HMM) and Conditional Random Field(CRF) are two famous models for supervised learning.

Shu et al.[35] used lexical HMM to train their model to learn patterns and to extract opinion terms and aspects.

Lei Shu et al. [36] showed that if a model has performed aspect extraction for the past times and has stored its knowledge this paired with CRF can boost accuracy from 81.3% to 84.3%

**Unsupervised:**

The actual word for this approach is topic modeling, which deals with the extraction of topics from texts, assuming that text does have a mixture of topics and every topic has a probability distribution over it. Probabilistic Latent Semantic Analysis(PLSA) and Latent Dirichlet Allocation(LDA) are two common methods under this approach. PLSA[37] is a novel method for the analysis of two-mode and co-occurrence data. LDA [38]is a method of generating a topic based on the number of occurrences in the set of documents.

In 2010, Zhao et al. [39] proposed a method known as Maximum Entropy LDA, or MaxEnt-LDA, a hybrid model that computes aspect words and aspect-specific opinion words simultaneously

Joint Sentiment Topic (JST) model was proposed by Lin et al. [40], this model considers topics and sentiments together.

A disadvantage of topic modeling is that it requires a large amount of data for training.

## IV   MODELS FOR SENTIMENT EXTRACTION

This section is divided into two subsections, Machine Learning Models and Deep Learning Models, given ass follow:

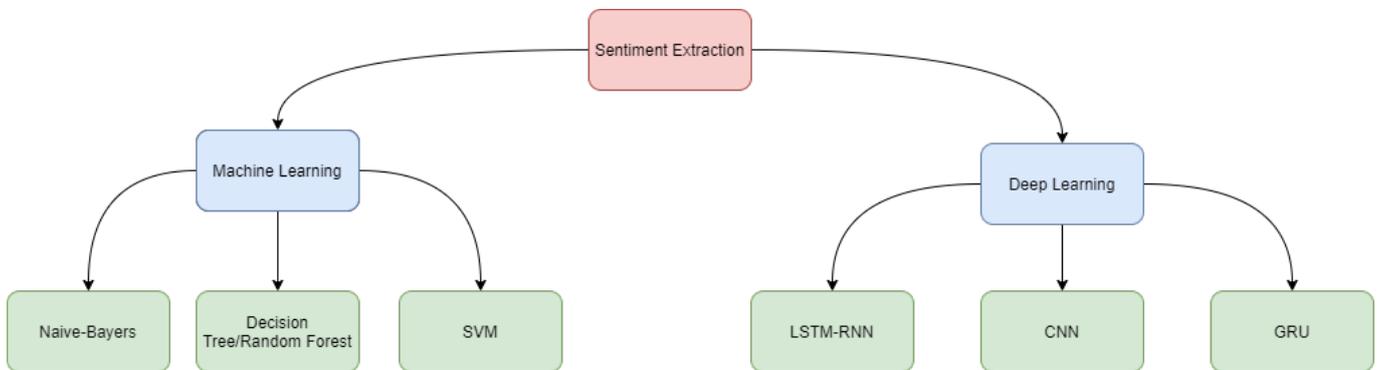

Figure 4: Classification of Sentiment Extraction on the different approaches used.

1. *Machine Learning Models*

**Naive Bayes:**

The paper we discuss for this model is Mubarok et al. [41]. Naive Bayes is a classification Machine learning algorithm. The reason why Naive Bayes is famous as it works well in the case of categorical inputs rather than numerical values. In the aforementioned paper, the authors used SemEval 2014 Task 4 dataset about product reviews on laptops and restaurants. For testing purposes, they divided the restaurant dataset as 97% and 3%, with the prior being the training set and the latter being the testing set.

The classification given by them was performed for two components, aspects and sentiments. Hence the probability distribution depends on both the components. They provided a conditional generative model that is explained as follows:

Let the document be d, and the set of sentiments is represented by S, and the set of aspects is A, represented as $(s_1, s_2, s_3, s_i)$ and $(a_1, a_2, a_3, a_i)$ and the document contains both the aspect and the sentiment which are represented as $(w_1, w_2, w_3, w_i, v_1, v_2, v_3, v_i)$ respectively. Now we have to calculate the probability of the document to be categorized into aspects and sentiments. This equation can be given by the probability of the aspects multiplied with the probability of aspect in the document(w) given the set of aspects, and the probability of the sentiment multiplied with the probability of sentiment in the document(v) given the set of sentiments. The equation can be represented by:

$$P(a_i, s_j | d) \propto \left( P(a_i) \prod_{1 \leq k \leq n_d} P(w_k | a_i) \right) \left( P(s_j) \prod_{1 \leq t \leq n_d} P(v_t | s_j) \right)$$

Now to calculate the final sentiment associated with the aspect the authors used Maximum A Posteriori(MAP) rule.

Hence the final sentiment is calculated using the following equation.

$$c_{MAP} = \arg\max_{a_i \in A, s_j \in S} P(a_i, s_j | d)$$

**Decision Tree and Random Forest Classifier:**

The Decision Tree model is also a classification model that tries to predict the category of an entry. The way it works, for sentiment classification, is described as follows. For example, we are looking to classify, into two sentiments, positive and negative. The decision tree model learns by looking at the annotated database and make a hypothetical tree structure:

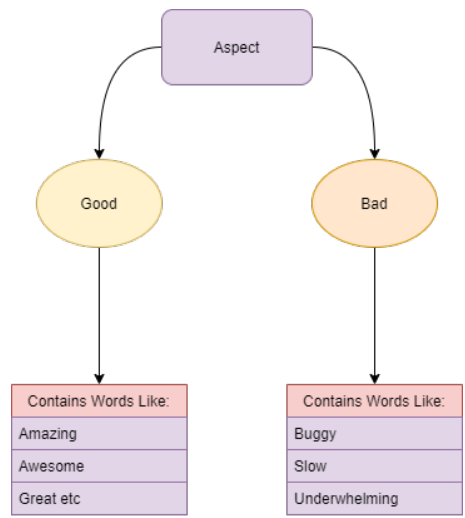

Figure 5: Decision Tree For a sentiment towards an Aspect

Now this tree is only for 2 sentiments but a Decision tree can have n-number of sentiments to it. The aforementioned model was improvised as there is very little text explaining the process. Some existing Models are, given in Asif et al.[42] and Jeevanandam Jotheeswaran [43]. Also, Random Forest Classifier is just aggregation of multiple decision trees, one thing to keep in mind is that this method can lead to over-fitting.

**Support Vector Machines:**

Support Vector Machines is also a very extensively used and a powerful classification tool, in the paper, we discuss variations of SVM are provided. To know more about SVM itself we refer the reader to Burges et al. [44], and for text classification, Joachims et al.

To give an SVM approach we selected Tony et al.[45], for pre-processing they used semantic orientation using PMI, which is described in the above sections. They used a movie review dataset that had a total of 1380 entries, approximately half good and half bad. The same dataset was used by Pang et al.[46], and the authors to compare results used the same 3 -fold cross-validation as Pang et al., and for more conclusive results the authors also give a 10-fold cross-validation experiment. The second dataset used by them was only a 100 records long dataset, so they provided 20,10,5 fold cross-validations, it's important to note that their experiments were totally focused on SVM itself, not its variations.

The first variation we will discuss is Gini-SVM Chakrabartty[47]. It's the aggregation of two separate components, the Gini-Simpson Index and SVM. The Gini-Simpson Index is a biological concept that was used to calculate the diversity in a population and was given by the following formula:

$$1 - \sum_{i=1}^{k} \frac{n_i(n_i - 1)}{n(n-1)}$$

But in this context, we need the Gini-Simpson Maximum Entropy given by:

$$Entropy = -\sum_{j} p_j \log_2 p_j$$

So the Gini-SVM combines this quadratic entropy with a kernel-based model. The GiniSVM normalizes classification margins and directly gives the conditional probabilities, which is less computationally intensive assuming we approximate the Kernel Logistic Regression(KLR), as compared with other SVM models Gini-SVM produces a sparse kernel expansion.

For the next model, we will talk about is the Feature-Based SVM [48], instead of being a sentiment extraction it is more a trick to show the importance of a feature. The authors used classifier coefficients. These coefficients were then used orthogonal vector coordinates that are orthogonal to the SVM hyperplane. Hence the feature importance can be determined by comparing the value of these coefficients. And by using this approach we can remove the un-important features which have less variance.

And lastly, we will discuss the Lexicon-Based SVM[49]. The approach to this method is quite simple. There is a defined set of rules according to ABSA is achieved.

Moving one more step is the complete rule-based approach, given by Omar et al.[50]. The first is the basic Lexicon(L) which is considered a baseline for the experiments. Then other rule settings were introduced such as handling negations, intensifiers("Absolutely useless"," absolutely"), downtoners("Pretty good but buggy"," pretty"), repeated characters("gooood") and special cases.

*2. Deep Learning Models*

**Convolution Neural Networks and GRU:**

Convolution Neural Networks can be broken down into two phrases a convolution and Neural Network. A Neural Network is defined as the set of artificial neurons that behave like biological neurons in some sense. Whereas a convolution is referred to as a filter that is applied to a kernel to get a feature map. To study the application of CNN's in ABSA we selected two papers that give an in-depth look into the concepts.

The first paper we selected uses basic CNN, Dwi et al.[51], their model can visually be represented as follows.

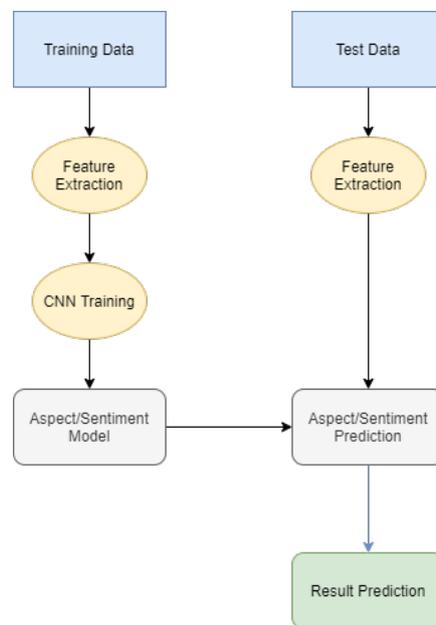

Figure 6: Generic Procedure for CNN Models

Although this paper does dive into aspect and feature extraction we will focus more on CNN training, but for Feature the authors used Glove, word-embeddings. The CNN receives input in the form of vector representations. Hence the first order of work is generating split vector representations for each vector with its label. CNN training can also be improved by hyper-tuning the parameters so as to get an optimized result.

The architecture inputs 100*100 vector representations and the output can give multiclass outputs for aspect modeling or binary classification for sentiment modeling. The model involves the use of 2 layered convolutions for the filter and the layer itself has 3 regions and 2 filters. Convolution has its own activation, but the layers in the convolution process use the same activation. The activation used is ReLU which produces the same value as the input but gives 0 when the input is less than 0. After this comes max-pooling which gives a smaller vector value. Then all vectors are combined into one vector. Then comes final activation that generates a representation of each class attribute. The activation is used as a softmax or a sigmoid depending on the requirements.

This process is used as a baseline, we can move onto parameters such as filter size, Max pool Size, Drop Out, Epoch, etc. for optimized results for different datasets.

The other paper we review uses Gated CNN, instead of the generic CNN. The gating causes the gradient to flow back without downscaling or vanishing through a linear path. This is called the Gated Linear Unit( GLU)  and it also called multiplicative skip connection.

The model first takes the Context Embedding and the target embeddings and passes them through a convolution layer separately. Then the max-pooling of the target embedding is calculated and passed with the convolutional result of the Context Embeddings, not the GLU is used here to get faster results, and then the results from every layer are max pooled to get a resultant vector. Now, this vector is then passed through Adaptive Softmax[53] to get the final sentiment. The reason for using softmax is to handle big datasets.

*LSTM-RNN:*

Long Short Term Memory or LSTM is a very extensively used technique in the field of ABSA, although it's old it is still being researched. One of the problems with generic RNN was backpropagation and gradient vanishing problem. What this means is that as we move on to the successive layers of an RNN, the gradient vanishes as we backpropagate. For example, there is a paragraph that starts with "I'm from Germany, and I can speak german" and then is followed by 1000 lines and then ends with "I can fluently speak__" and we have to predict the underlined content. The RNN should easily predict this as it is mentioned at the start of the document and it can easily backpropagate to get the answer, but this is the problem with RNN, the gradient reduces as we move a layer back, vanishes as we backtrack to the initial layers. So over the long-term, the RNN forgets easily. There are two factors that affect the gradient: weights and the activation functions. To solve this LSTM [55] cell was introduced. The difference between RNN and LSTM is that it has 3 gates, input, forget and an output gate. So as compared to the RNN there are more matrix computations. And an LSTM cell is fully differentiable that is there is individual weighted matrix associated with the gates. So the input gate is fed data, the model first forgets any long-term data that it decides is not needed anymore and then it decides which parts of the new input are important to be remembered for long. So instead of learning everything, it focusses more on the parts that are important. The equations of every state are as follows:

$$f_i = \sigma(W_f[x_i, h_{i-1}] + b_f)$$
$$I_i = \sigma(W_I[x_i, h_{i-1}] + b_I)$$
$$\tilde{C}_i = tanh(W_C[x_i, h_{i-1}] + b_C)$$
$$C_i = f_i * C_{i-1} + I_i * \tilde{C}_i$$
$$o_i = \sigma(W_o[x_i, h_{i-1}] + b_o)$$
$$h_i = o_i * tanh(C_i)$$

where $f_i$, $I_i$, and $o_i$ are the forget gate, input gate and the output gate. $W_f$, $W_i$, $W_o$, $b_f$, $b_i$, and $b_o$ are weight matrix and bias scalar for each gate $C_i$ is the cell state and $h_i$ is the hidden output.

In this section, we will discuss a novel method introduced by Yukun et al. [54]. They introduced a new variation of LSTM ie. a Sentic LSTM. Their architecture contained a sequence encoder and a hierarchical attention component.

The work-flow is discussed as follows:

A sentence is entered and a lookup operation is performed converting the input words into embeddings. Then the sequence-encoder converts then converts the word embeddings to a sequence of hidden outputs. Now, the attention component is built on top of these hidden outputs. the target-level attention takes these hidden outputs at the target expression positions and calculates self-attention over these. This output is then with the word embedding is used to get the sentence -level attention converting the whole sentence into a vector. The sentence-level attention will give one sentence vector for each aspect, this is then fed to the multiclass of aspects (Neutral, Negative, Positive, None).

## V    EVALUATION

Now, we move onto the most important part of this study that is the comparison of the above-mentioned models. This section is also divided into 3 sections: pre-processing, Machine Learning and Deep Learning. It is vital to note that all models don't have a unified metric system or the datasets used. Hence, we have to completely rely on the data given by the individual papers. Also, we have added other models as well, which are the derivatives or similar to the models we provide.

We will use 3 different metrics for our comparison, F1 scores, Precision, and Recall. After preprocessing, model section, and application comes evaluation, which actually shows how good did your model performs. Firstly we will give the meaning and the requirements for the 3 metrics and then show the results.

Before we move to the metrics we should know what true positive, true negative, false negative and false positive are.

> **True Positives (TP):** The case in which the actual value of a class was 1 and the predicted class was also 1
> **True Negatives (TN):** The case in which the actual value of a class was 0 and the predicted class was also 0
> **False Negatives (FN):** The case in which the actual value of a class was 1 and the predicted class was 0
> **False Positives (FP):** The case in which the actual value of a class was 0 and the predicted class was 0

The ideal situation would be to only get TP and TN, but that is not the case in most of the situations.

The metrics that are most widely used for evaluation are described as follows:

> **Precision:** Precision is defined as the number of true positives over the sum of true positives and false positives.
> **F1 Score:** F1 Score is defined twice the product of Precision and Recall over the sum of Precision and Recall
> **Recall:** Recall is defined as the number of true positives over the sum of true positives and false negatives.
> **Specificity:** Specificity is defined as the number of true negatives over the sum of false positives and true negatives.
> **Confusion Matrix:** It is a matrix with the number of true positive, true negative, false negative and false positive

| Model Name | Dataset Used | Precision | F1-Score | Recall |
|---|---|---|---|---|
| Dependency Parser | Amazon/C.net/Nintendo Review Data | 0.483 | 0.529 | 0.585 |
| Double Propagation | Amazon/C.net/Nintendo Review Data | 0.88 | 0.86 | 0.83 |
| POS-Thresholding | PASCAL VOC 2007 | 0.642 | 0.693 | - |
| PMI | Amazon/C.net/Nintendo Review Data | 0.72 | - | 0.91 |
| TF-IDF + SVM | Nintendo Review Data | 0.924 | 0.960 | 1.00 |
| CRF | Amazon/C.net/Nintendo Review Data | 0.843 | 0.773 | 0.714 |
| HMM | Amazon Review Data | 0.831 | 0.808 | 0.819 |
| LDA | EachMovie Colaborative Dataset | 0.95 | - | - |
| ASUM | SemEval 2016 Task 6 | 0.850 | - | - |
| JST | IMDb Movie Dataset | 0.82 | - | - |

Table 1: Evaluation for Methods for Aspect Extraction

| Model Name | Dataset Used | Precision | F1-Score | Recall |
|---|---|---|---|---|
| Naive-Bayes | SemEval 2014 Task 4 | 0.781 | 0.781 | 0.781 |
| Decision Tree/Random Forest | IMDb Movie Dataset | 0.576 | - | 0.661 |
| SVM | SemEval 2016 Task 6 | - | 0.750 | - |
| SVM+Lexicons | SemEval 2014 Task 4 | - | 0.829 | - |
| Feature-Based SVM | SemEval 2014 Task 4 | 0.66 | 0.64 | 0.61 |
| Gini-Index SVM | SemEval 2014 Task 4 | 0.80 | 0.75 | 0.779 |
| L+N+I+D+R+S | SemEval 2014 Task 4 | 0.92 | 0.64 | 0.79 |
| Lexicon Based SVM | SemEval 2014 Task 4 | 0.84 | 0.77 | 0.72 |
| CNN | SemEval 2016 Task 6 | 0.84 | - | - |
| Gated-CNN | SemEval 2016 Task 6 | 0.85 | - | - |
| GLU | SemEval 2016 Task 6 | 0.847 | - | - |
| Sentic LSTM-RNN | Senti-Hood | - | 0.764 | - |

Table 2: Evaluations of Sentiment Extraction Models

Table 1 refers to the performance of the Aspect Extraction part of the process, we can see that the Frequency-Based Approach is the most successful and TD-IDF is the most powerful as it gives a 92.4% precision. But there are limitations to the frequency-based approach, that is that it requires a huge data set and isn't really variable for Natural Language.

Table 2 shows the evaluations from the ABSA part of the process. As observed that a lot of entries are missing, this is due to the papers not providing all the metrics we chose to work on. If we look at the given set of values, we can see that the rule-based approach tanks all the other models, with a 92% precision but this model need a lot of defined set of rules and won't perform well if undefined instances occur. The next closet is the GCNN with an 85% precision.

## VI. CONCLUSION

As we have seen previously that ABSA is being used for making sense of ocean or review data, it is very crucial that every aspect amd it's sentiment is decoded. Hence we need systems that are accurate and also scalable for larger datasets. In the methods we discussed the best precision was shown by a rule based method, but as we discussed it's not scalable for other datasets. Newer models use both Machine/Deep learning and rules-based approach to solve the problem, but this doesn't mean that they absolutely will have accuracy or precision more than the models we have provided. A mixed approach can also negatively affect the results [56], as we can see the Rule Based and CNN give a precision of 0.7 where as normal CNN which we used in our research gave a precision of 0.84 and both the studies used the same dataset for their conclusions. Hence to get good results we need definition of rules and a model that works constructively with the rules provided.

REFERENCES

[1] P. Ray and A. Chakrabarti, A Mixed approach of Deep Learning method and Rule-Based method to improve Aspect Level Sentiment Analysis, Applied Computing and Informatics, https://doi.org/10.1016/j.aci.2019.02.002

[2] Gaillat, T., Stearns, B., McDermott, R., Sridhar, G., Zarrouk, M., & Davis, B. (2018, July). Implicit and explicit aspect extraction in financial microblogs. In Annual Meeting of the Association for Computational Linguistics). Association for Computational Linguistics.

[3] Poria, S., Cambria, E., Ku, L. W., Gui, C., & Gelbukh, A. (2014, August). A rule-based approach to aspect extraction from product reviews. In Proceedings of the second workshop on natural language processing for social media (SocialNLP) (pp. 28-37).

[4] Liu, Q., Liu, B., Zhang, Y., Kim, D. S., & Gao, Z. (2016, March). Improving opinion aspect extraction using semantic similarity and aspect associations. In Thirtieth AAAI Conference on Artificial Intelligence..

[5] Jihan, N., Senarath, Y., Tennekoon, D., Wickramarathne, M., & Ranathunga, S. (2017, November). Multi-Domain Aspect Extraction using Support Vector Machines. In Proceedings of the 29th Conference on Computational Linguistics and Speech Processing (ROCLING 2017) (pp. 308-322).

[6] Shu, L., Xu, H., & Liu, B. (2017). Lifelong learning crf for supervised aspect extraction. arXiv preprint arXiv:1705.00251.

[7] Silva, A. P., Silva, A., & Rodrigues, I. (2015). An approach to the POS tagging problem using genetic algorithms. In Computational Intelligence (pp. 3-17). Springer, Cham.

[8] H. Saif, H. Yulan, H. Alani, Semantic sentiment analysis of twitter, The Semantic Web-ISWC, Springer, 2012, pp. 508–524.

[9] E. Kouloumpis, T. Wilson, J. Moore, Twitter sentiment analysis: the good the bad and the omg!, ICWSM 11 (2011) 538–541

[10] Jaspreet Singh, Gurvinder Singh, Rajinder Singh, Prithvipal Singh, Morphological evaluation and sentiment analysis of Punjabi text using deep learning classification, J. King Saud University – Comput. Inform. Sci. (2018)

[11] Z. Feng, Jianxin R Jiao, Jessie Yang, L. Baiying, Augmenting feature model through customer preference mining by hybrid sentiment analysis, Expert Syst. Appl. 89 (2017) 306–317

[12] Wei Xue, Wubai Zhou, Tao Li, and Qing Wang. 2017. Mtna: A neural multi-task model for aspect category classification and aspect term extraction on restaurant reviews. In Proceedings of the Eighth International Joint Conference on Natural Language Processing (Volume 2: Short Papers), volume 2, pages 151–156

[13] Sepp Hochreiter and Jurgen Schmidhuber. 1997. ¨ Long Short-Term Memory. Neural computation, 9(8):1735–1780

[14] Dzmitry Bahdanau, Kyunghyun Cho, and Yoshua Bengio. 2014. Neural Machine Translation by Jointly Learning to Align and Translate. In ICLR, pages CoRR abs–1409.0473.

[15] Yequan Wang, Minlie Huang, Xiaoyan Zhu, and Li Zhao. 2016b. Attention-based LSTM for Aspect level Sentiment Classification. In EMNLP, pages 606–615.

[16] Meishan Zhang, Yue Zhang, and Duy-Tin Vo. 2016. Gated Neural Networks for Targeted Sentiment Analysis. In AAAI, pages 3087–3093.

[17] Wang, B., & Liu, M. (2015). Deep learning for aspect-based sentiment analysis. Stanford University report.

[18] Duyu Tang, Bing Qin, Xiaocheng Feng, and Ting Liu. 2016a. Effective LSTMs for Target-Dependent Sentiment Classification. In COLING, pages 3298– 3307.

[19] Chen, P., Sun, Z., Bing, L., & Yang, W. (2017, September). Recurrent attention network on memory for aspect sentiment analysis. In *Proceedings of the 2017 conference on empirical methods in natural language processing* (pp. 452-461).

[20] https://nlp.stanford.edu/software/tagger.shtml

[21] Manning, C. D. (2011, February). Part-of-speech tagging from 97% to 100%: is it time for some linguistics?. In International conference on intelligent text processing and

computational linguistics (pp. 171-189). Springer, Berlin, Heidelberg

[22] Rush, A. M., Reichart, R., Collins, M., & Globerson, A. (2012, July). Improved parsing and POS tagging using inter-sentence consistency constraints. In Proceedings of the 2012 joint conference on empirical methods in natural language processing and computational natural language learning (pp. 1434-1444). Association for Computational Linguistics

[23] Famili, A., Shen, W. M., Weber, R., & Simoudis, E. (1997). Data preprocessing and intelligent data analysis. Intelligent data analysis, 1(1), 3-23.

[24] Hatakoshi, Y., Neubig, G., Sakti, S., Toda, T., & Nakamura, S. (2014, October). Rule-based syntactic preprocessing for syntax-based machine translation. In Proceedings of SSST-8, Eighth Workshop on Syntax, Semantics and Structure in Statistical Translation (pp. 34-42).

[25] Angiani, G., Ferrari, L., Fontanini, T., Fornacciari, P., Iotti, E., Magliani, F., & Manicardi, S. (2016). A Comparison between Preprocessing Techniques for Sentiment Analysis in Twitter. In KDWeb.

[26] Qi, P., Dozat, T., Zhang, Y., & Manning, C. D. (2019). Universal dependency parsing from scratch. arXiv preprint arXiv:1901.10457.

[27] Zhuang, L., Jing, F., & Zhu, X. Y. (2006, November). Movie review mining and summarization. In Proceedings of the 15th ACM international conference on Information and knowledge management (pp. 43-50).

[28] Wu, Y., Zhang, Q., Huang, X., & Wu, L. (2009, August). Phrase dependency parsing for opinion mining. In Proceedings of the 2009 conference on empirical methods in natural language processing: Volume 3-volume 3 (pp. 1533-1541). Association for Computational Linguistics.

[29] Qiu, G., Liu, B., Bu, J., & Chen, C. (2011). Opinion word expansion and target extraction through double propagation. Computational linguistics, 37(1), 9-27.

[30] Smeaton, A. F., Kelledy, F., & O'Donnell, R. (1995). TREC-4 experiments at Dublin City University: Thresholding posting lists, query expansion with WordNet and POS tagging of Spanish. Harman [6], 373-389.

[31] Popescu, A. M., & Etzioni, O. (2007). Extracting product features and opinions from reviews. In *Natural language processing and text mining* (pp. 9-28). Springer, London.

[32] Blair-Goldensohn, Sasha, Kerry Hannan, Ryan McDonald, Tyler Neylon, George A. Reis, and Jeff Reynar, "Building a sentiment summarizer for local service reviews", In Proceedings of WWW-2008 workshop on NLP in the Information Explosion Era, 2008

[33] Shimada, K., Tadano, R., & Endo, T. (2011). Multi-aspects review summarization with objective information. *Procedia-Social and Behavioral Sciences*, *27*, 140-149.

[34] Moghaddam, Samaneh and Martin Ester, "Opinion digger: an unsupervised opinion miner from unstructured product reviews", In Proceeding of the ACM Conference on Information and Knowledge Management (CIKM-2010), 2010

[35] Shu, L., Liu, B., Xu, H., & Kim, A. (2016). Supervised Opinion Aspect Extraction by Exploiting Past Extraction Results. *arXiv preprint arXiv:1612.07940*.

[36] Shu, L., Xu, H., & Liu, B. (2017). Lifelong learning crf for supervised aspect extraction. *arXiv preprint arXiv:1705.00251*.

[37] Hofmann, T. (2013). Probabilistic latent semantic analysis. *arXiv preprint arXiv:1301.6705*.

[38] Blei, D. M., Ng, A. Y., & Jordan, M. I. (2003). Latent dirichlet allocation. *Journal of machine Learning research*, *3*(Jan), 993-1022.

[39] Zhao, Wayne Xin, Jing Jiang, Hongfei Yan, and Xiaoming Li, "Jointly modeling aspects and opinions with a MaxEntLDA hybrid", In Proceedings of Conference on Empirical Methods in Natural Language Processing (EMNLP-2010).

[40] Chenghua Lin, Yulan He, "Joint sentiment/topic model for sentiment analysis", In Proceedings of the 18th ACM conference on Information and knowledge management (CIKM '09), ACM, New York, NY, USA, pp. 375-384, 2009

[41] Mubarok, M. S., Adiwijaya, & Aldhi, M. D. (2017, August). Aspect-based sentiment analysis to review products using Naïve Bayes. In

*AIP Conference Proceedings* (Vol. 1867, No. 1, p. 020060). AIP Publishing LLC.

[42] Akhtar, M. S., Ekbal, A., & Bhattacharyya, P. (2016, April). Aspect based sentiment analysis: category detection and sentiment classification for Hindi. In *International Conference on Intelligent Text Processing and Computational Linguistics* (pp. 246-257). Springer, Cham.

[43] Jeevanandam Jotheeswaran, D. R., & Kumaraswamy, Y. S. (2013). Opinion mining using decision tree based feature selection through manhattan hierarchical cluster measure. *Journal of Theoretical and Applied Information Technology*, *58*(1), 72-80.

[44] Burges. 1998. A tutorial on support vector machines for pattern recognition. Data Mining and Knowledge Discovery, 2(2):121–167.

[45] Thorsten Joachims. 2001. Learning to Classify Text Using Support Vector Machines. Kluwer Academic Publishers.

[46] Mullen, T., & Collier, N. (2004, July). Sentiment analysis using support vector machines with diverse information sources. In *Proceedings of the 2004 conference on empirical methods in natural language processing* (pp. 412-418).

[47] Chakrabartty, S., & Cauwenberghs, G. (2007). Gini support vector machine: Quadratic entropy based robust multi-class probability regression. *Journal of Machine Learning Research*, *8*(Apr), 813-839.

[48] Krishna, M. H., Rahamathulla, K., & Akbar, A. (2017, March). A feature based approach for sentiment analysis using SVM and coreference resolution. In *2017 International Conference on Inventive Communication and Computational Technologies (ICICCT)* (pp. 397-399). IEEE.

[49] Ahmad, A. R., Viard-Gaudin, C., & Khalid, M. (2009, July). Lexicon-based word recognition using support vector machine and hidden markov model. In *2009 10th International Conference on Document Analysis and Recognition* (pp. 161-165). IEEE.

[50] Alqaryouti, O., Siyam, N., Monem, A. A., & Shaalan, K. (2019). Aspect-based sentiment analysis using smart government review data. *Applied Computing and Informatics*.

[51] Mulyo, B. M., & Widyantoro, D. H. (2018). Aspect-Based Sentiment Analysis Approach with CNN. *Proceeding of the Electrical Engineering Computer Science and Informatics*, *5*(5), 142-147.

[52] Xue, W., & Li, T. (2018). Aspect based sentiment analysis with gated convolutional networks. *arXiv preprint arXiv:1805.07043*.

[53] Grave, E., Joulin, A., Cissé, M., & Jégou, H. (2017, August). Efficient softmax approximation for GPUs. In *Proceedings of the 34th International Conference on Machine Learning-Volume 70* (pp. 1302-1310). JMLR. org.

[54] Ma, Y., Peng, H., & Cambria, E. (2018, April). Targeted aspect-based sentiment analysis via embedding commonsense knowledge into an attentive LSTM. In *Thirty-second AAAI conference on artificial intelligence*.

[55] Gers, F. A., Schmidhuber, J., & Cummins, F. (1999). Learning to forget: Continual prediction with LSTM.